\documentclass[a4paper,11pt]{article}
\usepackage[utf8]{inputenc}
\usepackage{amsmath,amssymb}
\usepackage{url}
\usepackage{indentfirst}
\usepackage[colorlinks,bookmarksopen,bookmarksnumbered,citecolor=blue,linkcolor=magenta]{hyperref}
\usepackage{graphicx}
\usepackage{gensymb}
\usepackage{xspace}
\usepackage{algorithm2e}

\newcommand{\sol}{\mathcal{S}}

\newcommand{\depo}{\pi}

\newcommand{\nentregas}{ne}
\newcommand{\eentregas}[1]{E[\nentregas(#1)]}
\newcommand{\size}{Length}
\newcommand{\peso}{w}
\newcommand{\Peso}{W}
\newcommand{\numeroRuas}{422\xspace}
\newcommand{\artur}{Artur Nogueira\xspace}

\newcommand{\nbb}{\mathbb{N}}
\newcommand{\qbb}{\mathbb{Q}}

\newcommand{\rota}{\mathcal{R}}
\newcommand{\rotamax}[1]{\rota_{max}(#1)}
\newcommand{\particao}[1]{Partition(#1)}

\newtheorem{definicao}{Definição}

\newcommand{\compri}[1]{W(#1)}

\newcommand{\atra}{Rg\xspace}
\newcommand{\atrana}{central\xspace}
\newcommand{\atranb}{peripheral\xspace}
\newcommand{\atranc}{distant\xspace}
\newcommand{\atrand}{isolated\xspace}

\newcommand{\atrb}{T}

\newcommand{\atrbna}{Avenue}
\newcommand{\atrbnb}{Street}
\newcommand{\atrbnc}{Alameda}
\newcommand{\atrbnd}{Highway}

\newcommand{\atrc}{Z}

\newcommand{\atrcna}{Commercial\&Industrial}
\newcommand{\atrcnb}{Mixed}
\newcommand{\atrcnc}{Residential}

\newcommand{\ruas}{Streets}

\newcommand{\dens}{D}

\newcommand{\penal}{Penal}

\newcommand{\tool}{VRPBench\xspace}

\title{\tool: A Vehicle Routing  Benchmark Tool} 

\newcommand{\footremember}[2]{%
    \footnote{#2}
    \newcounter{#1}
    \setcounter{#1}{\value{footnote}}%
}
\newcommand{\footrecall}[1]{%
    \footnotemark[\value{#1}]%
} 

\begin{document}

\maketitle

\author{Guilherme  A. Zeni\footremember{footFT}{School of Technology, University  of Campinas, Limeira -- SP -- Brazil }, \and Mauro Menzori\footrecall{footFT}, \and P. S. Martins\footrecall{footFT}, \and Luis A. A. Meira\footrecall{footFT}}

\vspace{8mm}
\begin{abstract}


The number of optimization techniques  in the combinatorial domain is large and diversified.
Nevertheless, there is still a lack of real benchmarks to validate  optimization algorithms.
In this work we introduce \tool, a tool  to create instances and visualize solutions  to the Vehicle Routing Problem (VRP)
in a planar graph embedded in the Euclidean 2D space. 
We use \tool to  model a real-world mail delivery case of the city of \artur. Such scenarios were characterized as a multi-objective optimization of the VRP.
We
 extracted a weighted graph from a digital map of the city to
create a challenging benchmark  for the VRP. 
Each instance models one generic day of mail delivery with hundreds to thousands of delivery points, thus
allowing both the comparison and validation of optimization algorithms for routing problems.

 \end{abstract}

\bigskip

\textbf{Keywords:} Benchmark, VRP, Graphs and Combinatorial Optimisation, Free Software Tool.

\vspace{8mm}

  \section{Introduction}
 \label{intro}
 
Benchmarks are found in various fields of science, such as geology~\cite{correia2015unisim}, economy~\cite{jorion1997value}, climatology~\cite{tol2002estimates}, among other areas. 
Specifically in computer science, benchmarks play a central role, e.g.  in image processing~\cite{du1990texture,krizhevsky2012imagenet,huang2007labeled}, hardware performance~\cite{che2009rodinia} and optimization~\cite{reinelt1991tsplib,kolisch1997psplib,burkard1997qaplib}.

In the context of optimization, David S. Johnson~\cite{johnson2002theoretician} divided  algorithm analysis in three approaches: the worst-case, the average-case, and the experimental analysis. Relative to experimental papers, he identifies four situations: \textit{(i)} to solve a real problem;
 \textit{(ii)} to provide evidence that an algorithm is superior than the others; \textit{(iii)} to better understand a problem; and \textit{(iv)} to study the average-case.
He suggests the use of well-established benchmarks  to provide evidence of the superiority of an algorithm (item \textit{ii}). Such papers are called 
\textit{horse race papers}.

Johnson highlights that  reproducibility and comparability are essential aspects present in any experimental paper. He also advocates the use of instances that lead to general conclusions.
The author mentions the difficulty in justifying experiments on problems with no direct application. Such problems have no real instances and the researcher is forced to generate the data in a \textit{vacuum}. Johnson cautions against a pitfall: 
the researcher starts by using randomly-generated instances to evaluate the algorithms and  ends up using the algorithm to explore the properties of the randomly-generated instances.
According to him, another pitfall is spending time processing useless experiments that attempt to answer  the wrong questions.

Our work deals with a variant of the Vehicle Routing Problem \textit{(VRP)} 
based on a real mail delivery case of the city of \artur (see the contributions at the end of this section).
 One of the first references to the VRP dates back to 1959~\cite{dantzig1959truck} under the name \emph{Truck Dispaching Problem}, a generalization of the Traveler Salesman Problem \textit{(TSP)}.

The term VRP was first seen in the paper by Christophides~\cite{christofides1976vehicle}, in 1976. Christophides defines VRP as a generic name, given to a class of problems that involves the visit of ``customers'' using vehicles.
 
Real world aspects may impose  variants of the problem.
For example, the \textit{Capacitaded-VRP (CVRP)}~\cite{fukasawa2006robust}
considers a limit to the vehicle capacity, the \textit{VRP with Time Windows (VRPTW)}~\cite{kallehauge2005vehicle}
accounts for delivery time windows, and the \textit{Multi-Depot VRP (MDVRP)}~\cite{renaud1996tabu} extends the number of depots.
  Other variants may be easily found in the literature.
 
In 1991, Reinelt~\cite{reinelt1991tsplib} created a benchmark for the TSP, known as TSPLib. In his work, he consolidated non-solved instances from twenty distinct papers. His repository  (TSPLIB95)~\cite{reinelt1995tsplib95} has instances of both the symmetric and the asymmetric traveling salesman problem (TSP/aTSP) as well as three related problems: \textit{(i)} CVRP; \textit{(ii)} Sequential Ordered Problem (SOP); and~\textit{(iii)} Hamiltonian Cycle Problem (HCP).

The number of instances is 113, 19, 16, 41, 9 for TSP, aTSP, CVRP, SOP, and HCP, respectively. The  number of vertices varies from 14 to 85900 for the TSP, 17 to 443 for the aTSP, 7 to 262 for the CVRP, 7 to 378 for the SOP, and from 1000 to 5000 for the HCP.


The optimum of all TSPLib instances was finally achieved in 2007,
after sixteen years of  notable progress in the development of algorithms.
The optimum of the  d15112 instance was found in 2001~\cite{applegate2011traveling}. This instance contains 15112 German cities and required  22.6 years of processing split across 110 500MHz processors~\cite{waterloosite}. 
The instance pla33810 was solved~\cite{applegate2011traveling} in March 2004. The pla33810 instance represents a printed circuit board with 33810 nodes and it was solved in 15.7 years of processing~\cite{espinoza2006linear}. 
The last instance of the TSPLib,  called pla85900, was solved in   2006~\cite{applegate2011traveling}. This instance contains  85900 nodes representing a VLSI application. 

Solomon~\cite{solomon1987algorithms} created a benchmark  for the  VRPTW in 1987. It is composed by 56  instances partitioned in 6 sets. 
The number of customers is 100 in all instances. The vehicle has a fixed capacity and the customers have demand or weight. The number of vehicles is not fixed, but it derives the fact that there is a limit of capacity. In this view, the problem can be considered multi-objective. It aims to minimize the route and the number of vehicles.

The first optimum solution was published in 1999~\cite{kohl19992}. In 2005, Chabrier~\cite{chabrier2006vehicle} managed to solve 17 of the instances that still had remained unsolved in the benchmark. In 2010, Amini et. al~\cite{amini2010pso} obtained solutions very close to optimum, considering the first 25 customers only. In july 2015, after 28 years of the benchmark launching, Jawarneh and Abdullah~\cite{jawarneh2015sequential} published in PlosONE a Bee Colony Optimization metaheuristic. Such algorithm reached 11 new best results in Solomon's 56 VRPTW 100 customer instances. It is really surprising that such small instances have an internal structure so complex to be optimized. In Figure 1, there is a simple instance of Solomon's composed by 100 customers and its solution considering 3 vehicles.
  
  \begin{figure}[ht]\centerline{
\includegraphics[width=.5\textwidth]{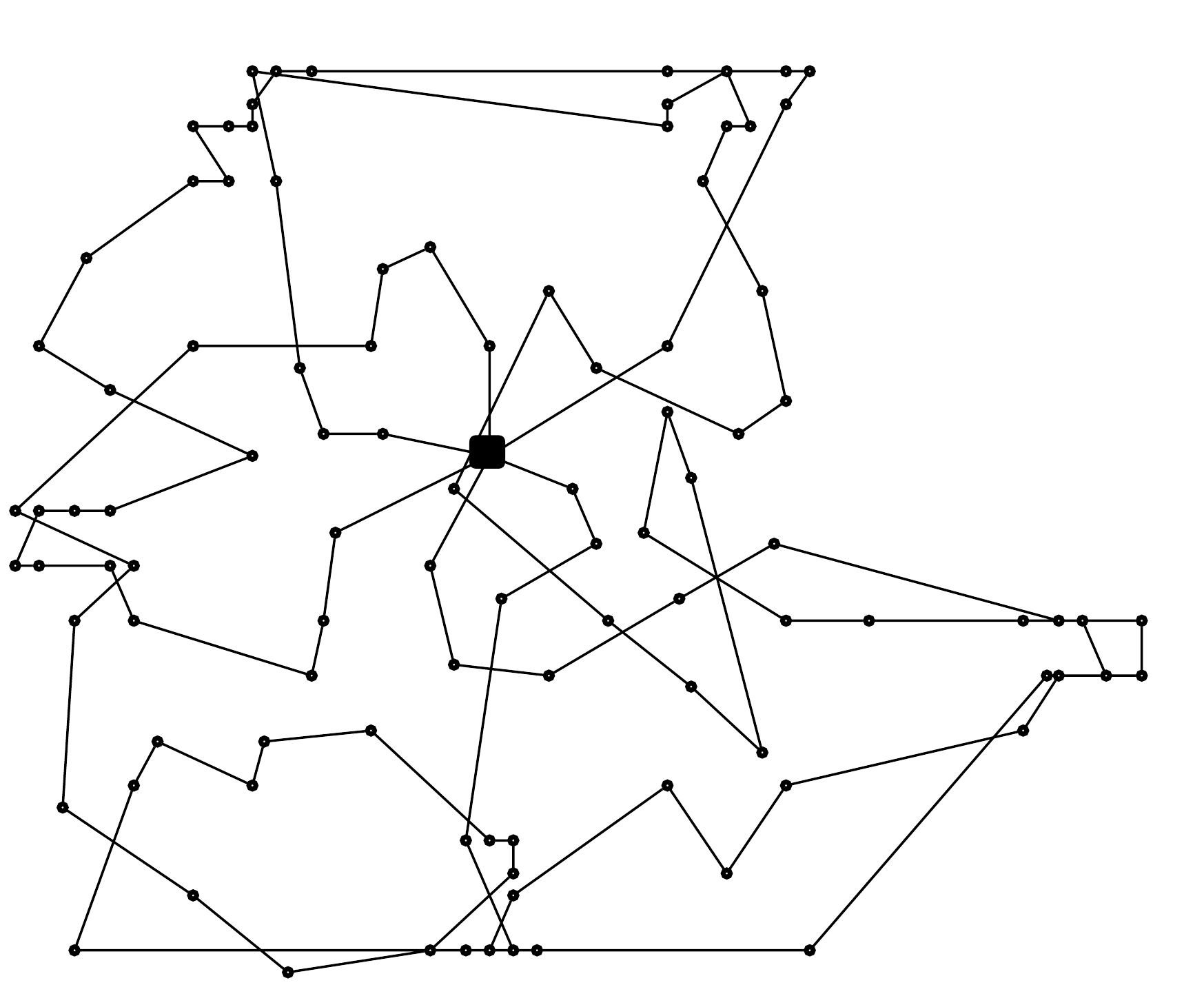}}
  \caption{The Solomon's RC207 instance composed by 100 customers and 1 depot. Bent and Hentenryck proposed the solution containing 3 routes, in 2001~\cite{bent2004two}. The picture does not show the capacity and the time windows constraints.}
  \label{figSolomonInst}
  \end{figure}
  
Despite of their complexity, the TSPLib and the Solomon benchmarks have a number of customers between 100 and 262 for the VRP, which is a small value nowadays. Gehring and Homberger~\cite{gehring1999parallel} extended the instances of Solomon, creating a benchmark with a numbers of customers varying from 100 to 1000 for the VRPTW.


For the CVRP, there is a set of instances largely used, named as ABEFMP, in which Augerat~\cite{augerat1995computational} proposed the classes A, B, P in 1995 and~\cite{christofides1969algorithm,fisher1994optimal,christofides1979vehicle} proposed the classes E, F, M in 1969, 1994 and 1979, respectively. In this benchmark, the number of customers varies from 13 to 200 and the number of vehicles varies from 2 to 17.

A series of different works, such as~\cite{fukasawa2006robust,contardo2014new} among others, obtained the optimum in different ABEFMP instances.~\cite{pecin2014improved} found in 2014 the optimum solution for the last instance unsolved, named \textit{M-n151-k12}, 35 years after its presentation by Christofides~\cite{christofides1979vehicle}. Despite that, most of those instances are very simple to solve nowadays.

Golden et al.~\cite{golden1998impact} proposed new instances for the CVRP, in 1998. It is a set of 20 instances, with the number of customers varying from 240 to 483. Such a benchmark remains entertaining, because most of its instances has not an optimum established yet~\cite{CVRPLIB}. In 2005, Li, Golden and Wasil~\cite{li2005very} created a set of instances with the number of customers between 560 and 1200. Up to this moment, there is no optimum defined for any of the instances~\cite{CVRPLIB}.

In 2014, Uchoa et. al. created a library, the CVRPLib~\cite{CVRPLIB}. In this library, they consolidated the CVRP instances of the works~\cite{augerat1995computational,christofides1969algorithm,christofides1979vehicle,fisher1994optimal,golden1998impact,li2005very}. Besides, Uchoa et. al~\cite{uchoanew} generated new instances with the number of customers between 100 and 1000. In this work, they point the lack of well established challenging benchmarks for the VRP.

Uchoa et. al. also point the fact that benchmarks are created artificially. Solomon and Uchoa et. al. generated their own instances using random points. In the benchmark ABEFMP, there is some random instances generated and other instances that represent real problems. However, in all instances the customers are points in the Euclidean space. The instances of Golden et al.~\cite{golden1998impact} and Li, Golden and Wasil~\cite{li2005very} are artificial as well.

\textbf{ Contribution:} 
{In this work, we propose the VRPBench tool,  a vehicle routing problem benchmark tool for the mail delivery Problem.}
We propose instances for the VRP involving a real situation: the delivery of correspondence by the postmen in the Artur Nogueira road network. Artur Nogueira is a brazilian city, located at $22\degree 34' 22" S 47\degree 10' 22" W$. Every day, the post office gets a large number (thousands) of letters to be delivered. The letters are distributed to a set of 20-25 postmen. The postmen walk around the city, delivering the mail.

The problem described above is modeled as VRP. Each postman is a vehicle and each delivery point is a customer. We abstracted the streets and avenues of the city in an weighted undirected graph. The street corners and the delivery points both became vertices and the streets became edges. The weight of an edge is proportional to the time needed to travel it.

After talking to experts, two objectives were detected to minimize: the average route length; and the injustice, measured as the unbalancing weight between the postmen.

The postmen do not have capacity, but the routes do have. Each postman can carry 10kg if man or 8kg if woman, however there is a car support that supplies the load to the postmen, which turns their capacities unlimited in mathematical terms. Meanwhile, each postmen must follow a working day close to 6h per day, which generates a maximum capacity for the route.

Defining capacity in the route length and not in the vehicle is something that applies to several situations. A helicopter has a route limited by its fuel tank. The workers, in general, have a time window to operate the vehicle, which limits the length of the route.

According to our knowledge, this is one of the first benchmarks for the VRP with thousands of customers and that models a real-life situation in a road network.


\section{Notation and Definition}

Consider a weighted (directed or undirected) graph $G(V,E)$  where 
a  $depot$ is a special vertice $\depo\in V$.  This research do not adresse multiple deposits variants of VRP .
Consider a cost function $w':E\rightarrow \qbb^+$

%

%
The set of customers is defined by  $C=V\setminus\{\depo\}$ and the number of customers is denoted by $n$, where $C=\{c_1,\ldots,c_n\}$.
%
The number of vehicles  in the fleet is represented by $k\in \nbb$. 
The value $k$ is traditionally considered a constant, but it is possible to define variants of VRP where $k$ is variable.

%
Let  $w(u,v,G)$  be the shortest path cost between vertices $u$ and $v$ in the graph $G$. We use  $w(u,v)$ to represent $w(u,v,G)$ whenever $G$ is known within the context.
%
We represent a solution as a sequence the vertices. Let
$$\sol(C,k) = (c_1,\ldots,c_n,\pi,\ldots,\pi).$$


This sequence is created as follows: \textit{(i)} all the customers are inserted
in  $\sol$; \textit{(ii)} the deposit vertice is inserted  $k-1$ times.

Each permutation of $\sol(C,k)$ represents a solution to the VRP. For example, consider the graph illustrated in Fig. \ref{fig:instance}:

\begin{figure}[htb]
\begin{center}
\includegraphics{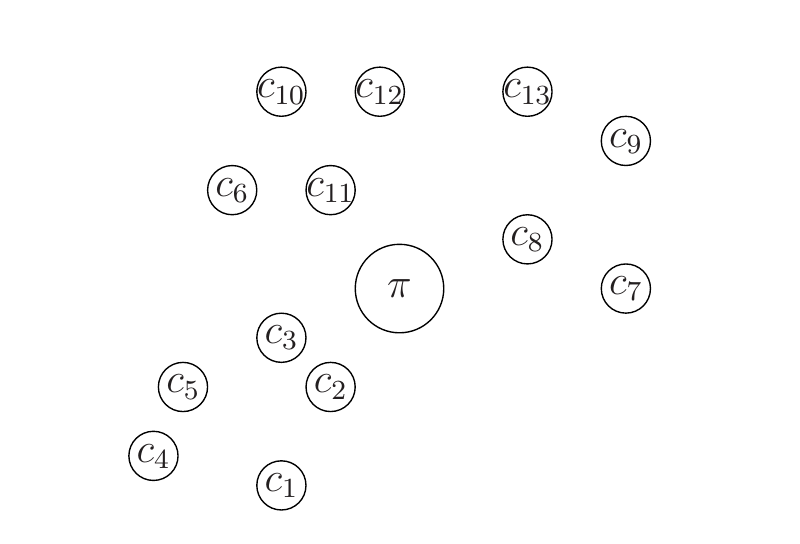}
\end{center}
\caption{Sample VRP Vertices.}
\label{fig:instance}
\end{figure}

%
Consider the vertices described in Fig. ~\ref{fig:instance} and suppose that the number of vehicles is 3 (i.e. $k=3$).
The  $\sol(C,3)$  sequence is: $$\sol(C,3)=(c_1,\ldots,c_{13},\pi,\pi).$$.

\begin{figure}[htb]
\begin{center}
\includegraphics{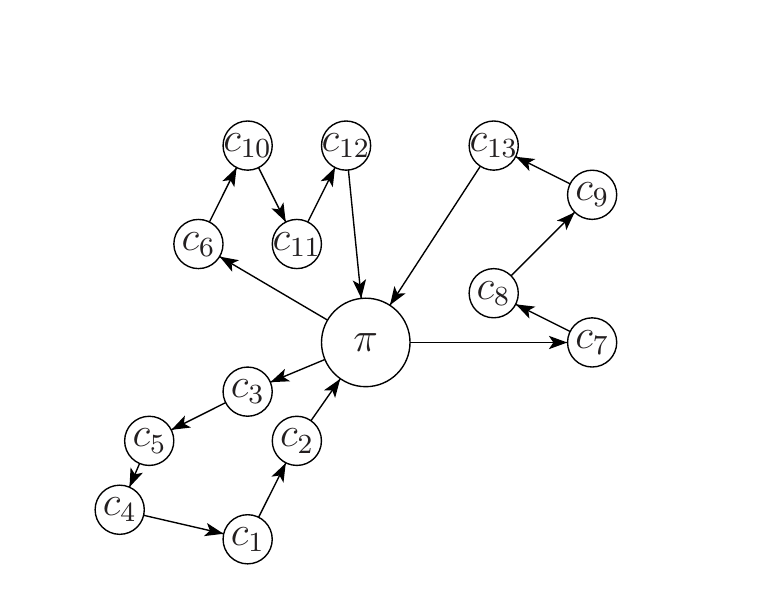}
\end{center}
\caption{Solution $\sol'=(c_3,c_5,c_4,c_1,c_2,\mathbf{\pi},c_6,c_{10},c_{11},c_{12},\mathbf{\pi},c_7,c_8,c_9,c_{13})$. }
\label{fig:solinst}
\end{figure}

Each permutation of $\sol(C,3)$ represents a solution to the VRP. For example, 
$$\sol'=(c_3,c_5,c_4,c_1,c_2,\pi,c_6,c_{10},c_{11},c_{12},\pi,c_7,c_8,c_9,c_{13})$$ 
is the solution described in Fig. ~\ref{fig:solinst}.

%
All routes begin and end at the depot.
The $\sol'$ solution represents a partition of the clients in 3 routes:  $\rota_1=(c_3,c_5,c_4,c_1,c_2)$, 
$\rota_2=(c_6,c_{10},c_{11},c_{12})$ and $\rota_3=(c_7,c_8,c_9,c_{13})$.
The vertex $\depo$ is used to create a partition of the sequence in $k$ routes: $$\displaystyle\particao{\sol}=\left(\rota_1,\ldots,\rota_k\right)$$ 
where the the sequence is split in each $k-1$ occurrence of   $\depo$.

%
%
In the example shown in Fig.~\ref{fig:solinst}.
we have  $$\particao{\sol'}=
\left(\left (c_3,c_5,c_4,c_1,c_2\right),\left(c_6,c_{10},c_{11},c_{12}\right),\left(c_7,c_8,c_9,c_{13}\right)\right)$$.

The length of a route $\rota=(r_1,\ldots,r_m)$  is given by:
$$\compri{\rota}=w(\depo,r_1)+w(r_m,\depo)+\sum_{i=1}^{m-1}w(r_i,r_{i+1}).$$
%
%
The length of a solution  $\sol=(s_1,\ldots,s_m)$ is calculated in as 
$$\compri{\sol}=w(\depo,s_1)+w(s_m,\depo)+\sum_{i=1}^{m-1}w(s_i,s_{i+1}).$$

Given a viable solution, it is necessary to calculate its  cost $f(\sol)$. 
The  most tradicional objective function to be minimized is the length of the
solution: $$f(\sol)=\compri{\sol}.$$.

However, in some situations the problem is considered as multi objective.
One of the objectives is to minimize $\compri{\sol}$ and the other 
is to minimize the number of vehicles.

$$f_1(\sol)=\compri{\sol},~~~~ f_2(\sol)=k.$$

%
Assume a system that generates mailman routes.
It is required that the solution meets the fairness criteria, i.e., routes are
assigned in a way that are not unfair to the mailman, i.e.
it does not penalize one mailman in relation to the other.
One way to achieve fairness would be to minimize the variance of the
length of the routes for each mailman.

As described in Section \ref{intro}, VRP is a set of problems that consist
in visiting customers using vehicles. Clearly, there are several variants of the VRP problem.
In each, the cost of the solution is calculated differently. Furthermore, each case
has additional requirements to find out whether or not a solution is viable.

Examples of restrictions to the problem are:

There should be no empty routes, i.e. for each route $\rota_i\in\particao{\sol}$ we must meet the
condition that $|\rota_i|>0$.

Each customer $c\in C$ has a demand $d(c)$ (e.g. number of orders).
Each  vehicle in $v\in\{1,\ldots,k\}$ is characterized by a capacity  $c(v)$. 
This variant of VRP models situations where a driver (e.g. mailman) 
has a maximum weight limit. Therefore, a route $\rota_v=(r^v_1,\ldots,r^v_m)$ 
must meet the requirement that $$\sum_{i=1}^md(r^v_i)\leq c(v),\forall v\in\{1,\ldots,k\}.$$.

Consider a variant of the VRP problem where the size of the route is limited.
This problem can model the scenario where vehicles need to
fuel at the depot (e.g. a helicopter). It also cater for
legal labour issues, where a driver has a commitment to minimum availability.
Let $v\in\{1,\ldots,k\}$ be a vehicle with a maximum route $\rotamax{v}$.  
Therefore, any given route $\rota_v=(r_1,\ldots,r_m)$ is constrained by the condition that
$$W(\rota_v)\leq \rotamax{v},\forall v\in\{1,\ldots,k\}.$$.

VRPTW is a variant of the VRP problem, where a visit to a vertice must be carried out within a time window.
In this case, the solution is deemed to be feasible if a client is within a pre-defined lower and upper time limit.
Examples of cases are scheduled deliveries such as SEDEX 10, which 
guarantees that the delivery is completed by the following day before 10 am.
It is also useful to implement  residential technical support, where the visit to a customer is
scheduled within a time frame.

A simple way to deal with unfeasible solutions is to assign an arbitrary high cost to the objective functions:
whenever a solution is considered unfeasible, its cost is infinite  ($f(\sol)\gets\infty.$).

For simplicity, in Section \ref{method} we will describe the incapacitated VRP with time windows. 
The number of vehicles is a constant and we seek to minimize the
total length of the routes ($\compri{\sol}$). We also assume there are
no empty or idle routes.

%
\begin{definicao}[Single-objective VRP] Given a weighted graph  $G(V,E)$, a constant  $k$, a special vertice
$\pi\in V$ and an objective function $f$. Seja $C\gets V\setminus\{\pi\}$. 
Consider the sequence $\sol(C,k)$ and let $P$ be the set of all permutations of $\sol(C,k)$. Find the permutation $\sol^*\in P$
so that  $f(\sol^*)$ is minimum.
\end{definicao}

\section{Methodology}
\label{method}

\begin{figure}[h!]
	\includegraphics[width=1.0\textwidth]{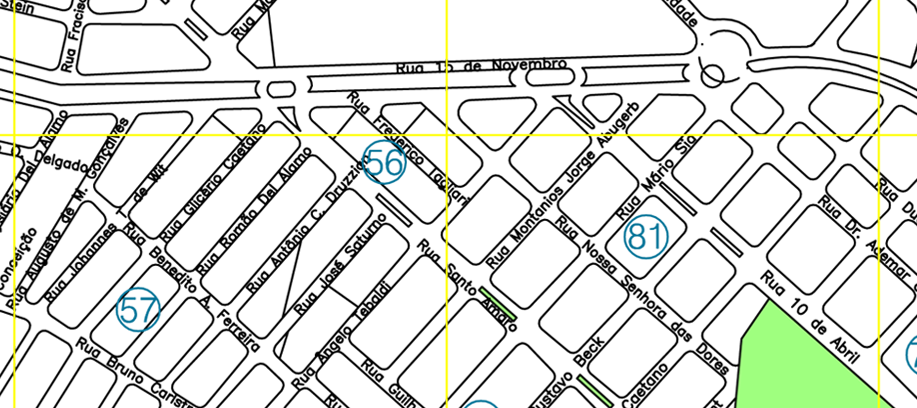}
  	\caption{A Section of the Map of  \artur.}
	\label{mapa}
\end{figure}

The instances were not generated from the actual post-office application data. Nevertheless, the first author 
has applied his domain expertise (having worked for four years in the field at a post office in Arthur Nogueira)
to turn the instances as realistic as possible. The starting point was the map of the city of Arthur Nogueira as shown in Fig.  \ref{mapa}.


\begin{figure}[h!]
	\includegraphics[width=1.0\textwidth]{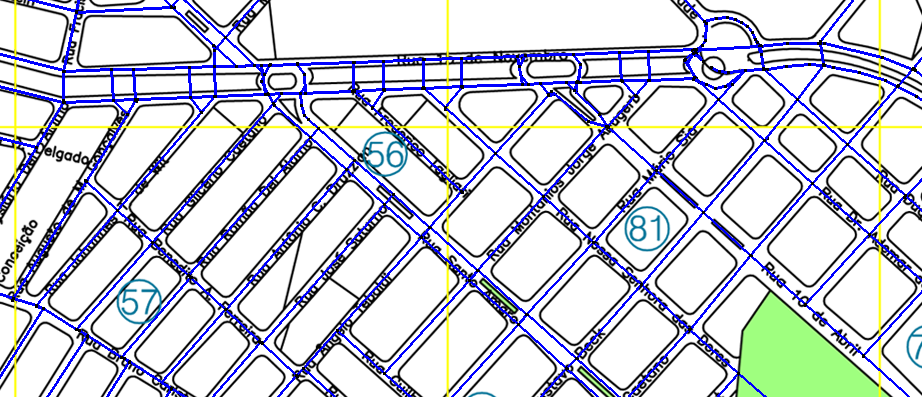}
  	\caption{Arestas e vértices criados sobre o mapa.}
	\label{mapa2}
\end{figure}

%
Line segments   representing the streets were drawn on top of the map (Fig. \ref{mapa2}).
Each corner was automatically identified by means of an algorithm that calculate intersections.
The result was a weighted graph where the weight of an edge represents the length of the segment.

Clearly, the edges representing deliveries that are carried out by a vehicle need to follow the  direction of one-way  streets. However, in our case,
postmen deliveries were performed by foot. 

We have created the graph directly from the street map due to the fact that:
 \textit{(i)} the path by foot may differ from the ones available from maps which prioritize delivery by vehicles;
 \textit{(ii)} the number of streets in \artur is slightly over 400 and that allows the manual creation of the graph;
 \textit{(iii)} Currently, public maps such as OpenStreetMap ~\cite{haklay2008openstreetmap}  are incomplete, i.e. they have a large number of
 streets not yet covered;

Ideally, the creation of new benchmarks for other cities should also consider public maps,
mainly in large cities where the manual extraction of the map may  incur in prohibitive costs.



The final graph ended up with $|V|=2111$ and $|E|=3225$. Each edge is associated with a pixel.
%
Each edge has a bijection with a line segment 
Each street has a name and a set of collinear line segments.
The cost of an edge is proportional to the length of its associated line segment, as discussed later.


An initial approach would be to consider the deliveries  generated according to a uniform distribution across the city streets.
However, a non-uniform distribution would be more realistic. For example, the main 
streets in the downtown area  do tend to receive a larger number of deliveries per unit length than the side or back streets.

Therefore, a relative density $\dens$ parameter was applied to each street. For example, a
street with  $\dens=.5$ has  a 50\% less probability of receiving a delivery than one with  $\dens=1.0$ (per unit length).




\subsection{Streets Density of Probabilty }
\label{sec:densi}

For each street, we arbitraly classify it with atributtes and levels.

There are tree attributes Region(R), Type(T) and Zone(Z). Both attributes and penalties were created based in
the expert knowledge.


The criterion by regions follows the premise that the downtown area closest streets have a higher rate of deliveries by length unit than the streets located at extreme regions of the city. 

We define a multiplicative penalty for each type of region. \\
\begin{tabular}{|c|c|c|c|c|c|}\hline
\textbf{Region} & \atrana &\atranb &\atranc &\atrand \\\hline
\textbf{Multiplic. Penalty~(\penal)} & 1.0 & 0.75 & 0.4 & 0.2 \\\hline
\end{tabular}
\bigskip

It means that, fixing other atttributes, an \atrand street receives 80\% less deliveries
than a \atrana street per unit of lenght.

We also define the attribute Type with four levels:

\begin{tabular}{|c|c|c|c|c|c|}\hline
\textbf{Type} & \atrbna &\atrbnb &\atrbnc &\atrbnd \\\hline
\textbf{Multiplic. Penalty~(\penal)} & 1.0 & 0.75 & 0.4 & 0.0 \\\hline
\end{tabular}
\bigskip

In this work, all highways received directly the value 0, because there were no delivers on it.

%

An street has 25\% menos entregas por unidade de comprimento do que uma avenida.

At last, we created the attribute Zone, which has three levels:
with penalties\\
\begin{tabular}{|c|c|c|c|c|c|}\hline
\textbf{Zone} & \atrcna &\atrcnb &\atrcnc \\\hline
\textbf{Multiplic. Penalty~(\penal)} & 1.0 & 0.7 & 0.4  \\\hline
\end{tabular}
\bigskip

We used the Google Maps as an auxiliary tool to verify the streets containing companies.

The criterion Zone exists because a street deliveries density does not depend only on its type or region. The economic context should also be regarded. 

%
Each one of the \numeroRuas streets received a value in $\atra\times\atrb\times\atrc$, according to expert knowledge.
Be $\ruas=(Street_1,Street_2,\ldots,Street_{\numeroRuas})$ the set of all  streets. Be $f:\ruas \rightarrow \atra\times\atrb\times\atrc$ a function that attributes values to a particular street. For instance, consider the $Street_{Rua XV}$. We have $f(Street_{Rua XV})$ was manualy define as $(\atrana,\atrbna,\atrcnb)$, because it is an avenue, spoted at downtown, containing residences and commercial establishments. 

 We can  refer to each $f$ element apart, in the following manner: $\atra(Street_{XV})=\atrana$, $\atrb(Street_{XV})=\atrbna$ and $\atrc(Street_{XV})=\atrcnb$.

Finally, we have a function that describes the relative density of a street. 

$D:\ruas\rightarrow \mathbb{N}$, as it follows:
$$D(Street)=\penal(\atra(Street))\times \penal(\atrb(Street))+\penal(\atrc(Street)).$$ 
Using this methodology, all the streets received a relative density.
For example, the XV de Novembro Avenue relative density is given by
\begin{eqnarray}
D(Street_{XV})&=&\penal(\atra(Street_{XV}))+\penal(\atrb(Street_{XV}))+\penal(\atrc(Street_{XV})) \nonumber \\
&=&\penal(\atrana)\times\penal(\atrbna)\times\penal(\atrcnb) \nonumber \\
&=&1\times 1\times 0.7=0.7 \nonumber \\
\end{eqnarray}

On the other hand, the Sibipirunas Alameda has the relative density:
\begin{eqnarray}
D(Street_{Sib.})&=&\penal(\atra(Street_{Sib.}))+\penal(\atrb(Street_{Sib.}))+\penal(\atrc(Street_{Sib.})) \nonumber \\
&=&\penal(\atrand)+\penal(\atrbnc)+\penal(\atrcnc) \nonumber \\
&=&0.2\times 0.2\times0.4 = 0.16 \nonumber \\
\end{eqnarray}

$D(Street_{XV})=0.7$ and $D(Street_{Sib.})=0.16$ are used so that, in our model, the probability density of the $Rua_{XV}$ are be $\frac{0.7}{0.14}$ times larger than probability density  $Street_{Sib.}$.

\subsection{Generating the Delivery Points}

We shall define a variable weight $w$ as
$$ \peso(Street) = D(Street)\size(Street).$$ 
So $w$ is directly proportional to the relative probability density and to the street length.


$$ \eentregas{Street} = \peso(Street)k,$$

Take a sequence of weights $(\peso_1,\peso_2,\ldots,\peso_n)$ associated to the streets
$(Street_1,Street_2,$
$\ldots,Street_n)$.

Thus, we will define a variable $\Peso$ as

$$
\Peso=\sum_{i=1}^{n}\peso(Street_i)
$$

The algorithm considers all the weights projected in a range $[0,W]$. 
For each delivery point, a random value $R\in[0,W]$ is created. Case this random value falls on the range of a particular street, the delivery 
point will be inserted in the respective street.

\begin{algorithm}[htb!]
\LinesNumbered
\SetAlgoLined
\KwIn{A sequence of weights $(\peso_1,\peso_2,\ldots,\peso_n)\in \mathbb{R}^n$ associated to streets and the total 
number of delivery $m\in\mathbb{N}$.}
\KwOut{The number of delivery for each street $(\nentregas_1,\nentregas_2,\ldots,\nentregas_n)\in \mathbb{N}^n$}
Set~$\displaystyle\Peso=\sum_{i=1}^{n}\peso(Street_i)
$\\
\For{i=1 to m}{
$R \gets$ random value int [0,W]\\
\For{x=1 to m}{
\If{$\displaystyle\sum_{j=1}^{x-1}w_j<R\leq\sum_{j=1}^{x}w_j$}
{$\nentregas_x\gets\nentregas_x+1$}
}
}
{  
    \Return{$(\nentregas_1,\nentregas_2,\ldots,\nentregas_n)$}
}
		
\caption{Generating delivery points.}
\label{alg:calckp}
\end{algorithm}

In a street, the $\nentregas(Street)$ delivery points are uniformly distributed.

\section{Using \tool to Model Manhatan Streets}

\section{Result}

We generated 10 sets of instances with 10 instances for set, resulting 100 instances.
The delivery points number of each set  is  1000, 2000, 3000, 4000, 5000, 6000, 7000, 8000, 9000 and 10000, respectively.


The figure below shows the distribution of 1000 delivery points, for a given area of the city chosen randomly:

\begin{figure}[h!]
	\includegraphics[width=1.0\textwidth]{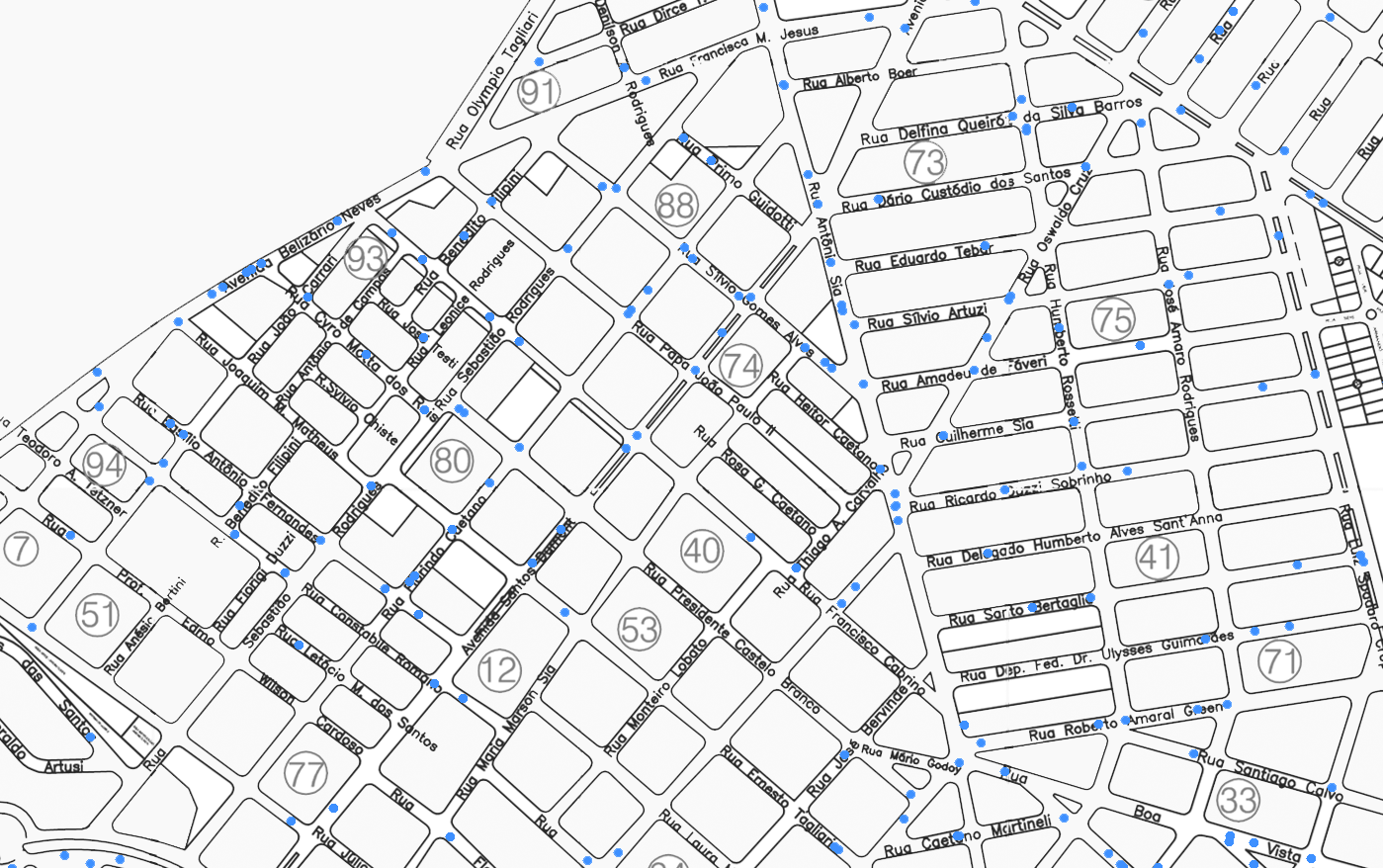}
  	\caption{Part of an instance generated with 1000 delivery points.}
\end{figure}

\section{Conclusions}


In this work, we created a benchmark based on a real-life situation. Once the solution used by post office is fixed with respect to the delivery points,
there is a large gap to be explored.

A feasible solution that reduces the number of postmen produces profit, since such postmen can be allocated in other tasks.

A solution that reduces the route length also reduce the delivery effort, and also can be understand as profit.

Furthermore, in science view, this is the first benchmark to VRP with 10.000 delivery points.

\subsection*{Future Perspectives}

From this work, we aim to publish a paper. We intend to develop a free software system to validate and to rank the best solutions found by the researchers, also providing a view of the routes. 
Lower bounds have a more complex validation and will be released only when associated to a publishing. 

At last, I would like to work in the development of an algorithm to solve some of the instances presented in this work, in a future master degree.

\bibliographystyle{plain}
\bibliography{tese}

\end{document}